\documentclass[10pt,twocolumn,letterpaper]{article}

\usepackage{cvpr}              

\usepackage{graphicx}
\usepackage{amsmath}
\usepackage{amssymb}
\usepackage{booktabs}
\usepackage{placeins}
\usepackage{float}
\usepackage{enumitem}
\usepackage{tabularx}
\usepackage{array}
\usepackage{ragged2e}
\usepackage{placeins}
\usepackage{afterpage}
\usepackage{caption}

\usepackage[pagebackref,breaklinks,colorlinks]{hyperref}

\usepackage[capitalize]{cleveref}
\crefname{section}{Sec.}{Secs.}
\Crefname{section}{Section}{Sections}
\Crefname{table}{Table}{Tables}
\crefname{table}{Tab.}{Tabs.}

\def\cvprPaperID{*}
\def\confName{CVPR}
\def\confYear{2025}

\begin{document}

\title{Inject Where It Matters: Training-Free Spatially-Adaptive\\Identity Preservation for Text-to-Image Personalization}

\author{Guandong Li\\
iFLYTEK\\
\quad (Corresponding Author)\\
\and
Mengxia Ye\\
Aegon THTF\\
}

\twocolumn[{
\renewcommand\twocolumn[1][]{#1}
\maketitle
}]

\begin{abstract}
Personalized text-to-image generation aims to seamlessly integrate specific human identities into arbitrary text descriptions. However, existing tuning-free methods (e.g., PuLID, InstantID) typically rely on \textbf{Spatially Uniform Visual Injection}. This ``global broadcast'' anchoring mechanism triggers a fundamental conflict between identity features and background context: strong identity features are injected not only into facial regions but also contaminate backgrounds, clothing, and environmental lighting, causing background semantic loss, style disconnection, or unnatural lighting when the target text involves specific scenes or styles. To break this dilemma without introducing expensive fine-tuning costs, we propose \textbf{SpatialID}, a novel training-free spatially-adaptive identity modulation framework. We fundamentally decouple identity injection into two spatial regions: \textbf{Face-Relevant Region} and \textbf{Context-Free Region}. Specifically, we design a \textbf{Spatial Mask Extractor} that leverages the output response of cross-attention to automatically locate the effective scope of identity features. Crucially, we propose a \textbf{Temporal-Spatial Scheduling} strategy. This mechanism simulates the dynamics of the generation process: using a \textbf{center Gaussian prior} to stabilize composition during the noise-dominated early stage; using \textbf{attention masks} to precisely anchor identity during the structure-forming mid stage; and \textbf{adaptively relaxing} mask constraints during the texture-refining late stage to allow natural environmental lighting penetration. Extensive experiments on the IBench benchmark demonstrate that SpatialID achieves SOTA-level performance in text adherence (CLIP-T: 0.281), visual consistency (CLIP-I: 0.827), and image quality (IQ: 0.523). Compared to existing methods, SpatialID significantly eliminates background contamination while maintaining robust identity features, greatly improving image quality and editability in complex scenes.
\end{abstract}

\section{Introduction}
\label{sec:intro}

In recent years, text-to-image generation (T2I) technology has achieved revolutionary progress~\cite{ho2020denoising,rombach2022high,podell2023sdxl,peebles2023scalable}, making it possible to generate high-quality, high-fidelity images from natural language descriptions. In this field, \textbf{Identity Customization} has become one of the most application-valuable directions. Its core goal is to preserve the identity features of a specific reference person during generation while allowing users to freely control the person's pose, expression, style, and environment through text prompts.

To achieve this goal, early fine-tuning-based methods (e.g., DreamBooth~\cite{ruiz2023dreambooth}, LoRA~\cite{hu2022lora}) performed well in identity preservation, but their expensive training costs and low deployment efficiency limited large-scale applications. Therefore, \textbf{Tuning-Free} identity injection methods have gradually become mainstream. These methods (e.g., PuLID~\cite{guo2024pulid}, IP-Adapter~\cite{ye2023ip}) typically use visual encoders to extract reference image features and inject them directly into the latent space of pretrained diffusion models through cross-attention mechanisms.

However, existing tuning-free methods face a long-overlooked challenge: \textbf{Spatially Uniform Visual Injection}. Most existing methods treat identity features as a set of globally broadcast signals, forcibly injecting reference image visual features with the same scalar weight regardless of whether an image region is a face, background wall, or sky:
\begin{equation}
    \mathbf{h} \leftarrow \mathbf{h} + \alpha \cdot \text{CA}(\mathbf{e}_{id}, \mathbf{h})
\end{equation}

This ``blind'' global anchoring mechanism leads to the \textbf{Identity Leakage} problem: skin color, textures, and even facial structures from the reference image are forcibly superimposed onto unrelated background regions. Specifically, when the prompt describes a particular environment (e.g., ``Mars surface'') or style (e.g., ``ink painting''), globally injected identity features often destroy the generation quality of the background, causing semantic conflicts and image quality degradation.

To address this dilemma, we revisit the role of cross-attention in generative models. We observe that although injection weights are global, the \textbf{Response Magnitude} of cross-attention naturally exhibits spatial sparsity---facial region responses are far higher than background responses. Based on this insight, we propose \textbf{SpatialID}, upgrading the scalar injection weight to a spatially-adaptive mask:
\begin{equation}
    \mathbf{h} \leftarrow \mathbf{h} + \alpha \cdot \mathbf{M}_t \odot \text{CA}(\mathbf{e}_{id}, \mathbf{h})
\end{equation}
where $\mathbf{M}_t \in [0,1]^{H \times W}$ is a spatial mask that varies with timestep $t$, and $\odot$ denotes element-wise multiplication. Unlike previous methods that attempt to balance conflicts through global parameter adjustment, SpatialID introduces a spatial decoupling paradigm of \textbf{``face region anchoring'' and ``background region release''}.

The core architecture of SpatialID consists of two key components:
\begin{enumerate}[leftmargin=*,nosep]
    \item \textbf{Spatial Mask Extractor}: Without introducing any additional detection models or parameters, we directly use the L2 norm of cross-attention outputs during generation to compute ``identity relevance'' in real-time. This enables the model to automatically perceive which regions need identity preservation and which regions should generate freely.
    \item \textbf{Temporal-Spatial Scheduling}: Considering the time-varying nature of the diffusion model generation process (from early noise composition to late texture refinement), static masks often fail. We design a three-stage scheduling strategy: using a \textbf{center Gaussian prior} to combat noise interference in the early stage; using \textbf{attention masks} to precisely lock the face in the mid stage; and allowing natural light-shadow fusion through a \textbf{mask relaxation} mechanism in the late stage, preventing artifacts at facial edges.
\end{enumerate}

In summary, the main contributions of this paper are as follows:
\begin{enumerate}[leftmargin=*,nosep]
    \item \textbf{Proposing the SpatialID framework}: The first tuning-free architecture that decouples identity injection into a spatially-adaptive process, fundamentally solving the background contamination and semantic conflict problems caused by global injection.
    \item \textbf{Temporal-Spatial Scheduling}: Designing a scheduling strategy that conforms to diffusion generation dynamics, dynamically adjusting the form and strength of spatial constraints at different denoising stages, without any additional training.
    \item \textbf{SOTA performance}: Achieving the best results in text alignment (CLIP-T: 0.281), visual consistency (CLIP-I: 0.827), and image quality (IQ: 0.523) on the IBench benchmark, surpassing 6 recent methods including PuLID, Dreamo, UNO, and DVI.
\end{enumerate}

\section{Related Work}
\label{sec:related}

\subsection{Personalized Text-to-Image Generation}

With the development of diffusion models~\cite{ho2020denoising,rombach2022high,song2020denoising,li2026flexidtrainingfreeflexibleidentity}, personalized generation has become a focus of the community. Early mainstream methods primarily relied on test-time fine-tuning. For example, Textual Inversion~\cite{gal2022image} optimizes specific text word vectors to represent new concepts; DreamBooth~\cite{ruiz2023dreambooth} implants subjects by fine-tuning denoising network weights; LoRA~\cite{hu2022lora} reduces the number of fine-tuning parameters through low-rank adaptation. Although these methods are effective, the time-consuming training for each ID limits their real-time application potential. This has driven the research community to explore more efficient tuning-free solutions, the domain where SpatialID operates.

\subsection{Tuning-Free Identity Preserving Generation}

The core idea of tuning-free methods is to use pretrained encoders (e.g., ArcFace~\cite{deng2019arcface}, CLIP~\cite{radford2021learning}) to extract features and inject them into the model. IP-Adapter~\cite{ye2023ip} introduces image prompts through a decoupled cross-attention mechanism. InstantID~\cite{wang2024instantid} uses ControlNet~\cite{zhang2023adding} combined with ID embeddings and facial landmarks for strong control. PuLID~\cite{guo2024pulid} achieves extremely high-fidelity injection through a ``lightning branch'' and contrastive alignment loss, using IDFormer to fuse ArcFace and EVA-CLIP~\cite{sun2023eva} features into 32 identity tokens, injecting through PerceiverAttention cross-attention at 20 injection points in the FLUX DiT~\cite{peebles2023scalable} architecture. PhotoMaker~\cite{li2024photomaker} adopts stacked ID embeddings. FastComposer~\cite{xiao2025fastcomposer} achieves tuning-free multi-subject generation with localized attention. Arc2Face~\cite{papantoniou2024arc2face} builds a foundation model for ID-consistent face generation. EditID~\cite{li2025editid,li2025editidv2} proposes training-free editable ID customization with data-lubricated feature integration. Dreamo~\cite{mou2025dreamo} and UNO~\cite{wu2024uno} explore unified approaches for multi-concept personalization, while UMO~\cite{cheng2025umo} scales multi-identity consistency via matching reward. However, most of the above methods adopt a \textbf{globally uniform injection strategy}, broadcasting identity features at the same strength across the entire image. SpatialID points out that this strategy is the main cause of background texture collapse and text adherence degradation, and proposes a spatially-adaptive solution.

\subsection{Spatial Control and Attention Mechanisms}

To improve generation controllability, Attend-and-Excite~\cite{chefer2023attend} and similar works enhance text alignment by manipulating cross-attention maps; ControlNet~\cite{zhang2023adding} and T2I-Adapter~\cite{mou2024t2i} introduce spatial conditions through additional encoders; OminiControl~\cite{tan2025ominicontrol,tan2025ominicontrol2} provides minimal universal control for diffusion Transformers. In the domain of spatial-aware generation, Li~\cite{li2024layout} proposes attention loss backward for layout control and semantic guidance, while mask-guided inpainting~\cite{li2024commerce} demonstrates the effectiveness of spatial masks in controllable generation. DVI~\cite{li2025dvi} attempts to decouple identity and attributes by learning independent semantic and visual representations, but still applies injection globally. The innovation of SpatialID lies in our first use of the \textbf{Response Map} of cross-attention itself as a self-supervised signal, achieving spatial decoupling of identity injection without external masks (e.g., Face Parsing).

\section{Method}
\label{sec:method}

SpatialID aims to solve the global injection contamination problem in tuning-free identity customization. We design the overall architecture as \textbf{Attention-Based Spatially-Adaptive Modulation}.

\begin{figure*}[t]
\centering
\includegraphics[width=\textwidth]{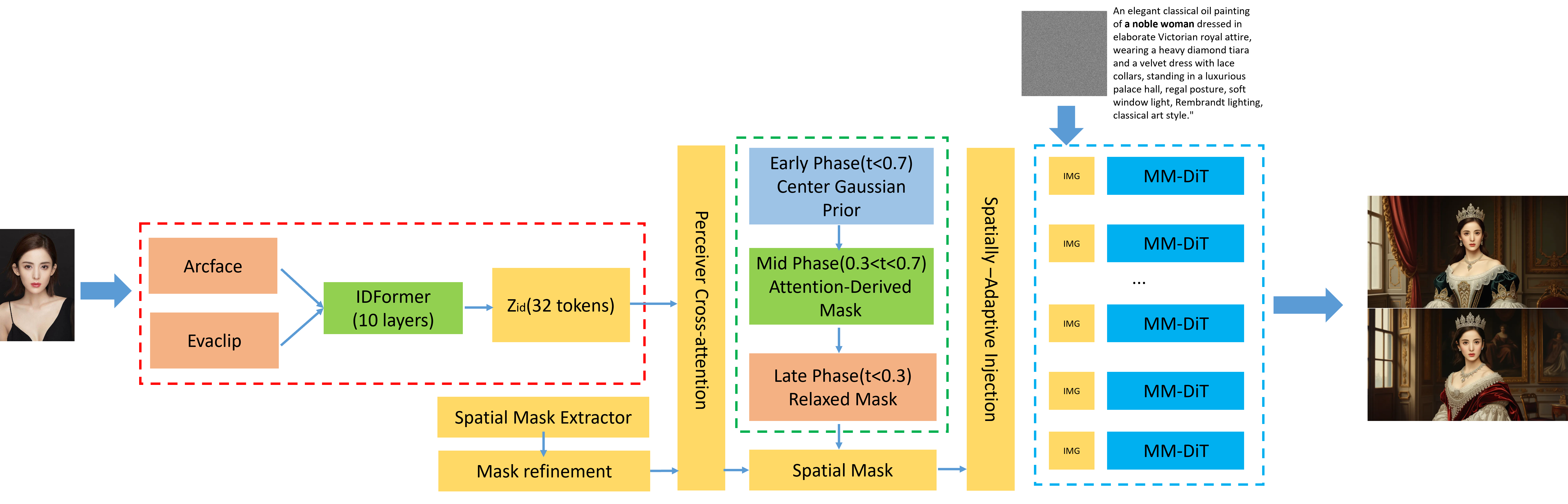}
\caption{Overall architecture of SpatialID. The reference face image is processed by the PuLID ID Encoder (ArcFace + EVA-CLIP + IDFormer) to extract identity tokens $\mathbf{Z}_{id}$. At each injection point in the FLUX DiT, the SpatialID module extracts a spatial relevance mask $\mathbf{M}_t$ from the cross-attention output, and dynamically adjusts the mask form across different denoising stages through the Temporal-Spatial Scheduler, achieving spatially-adaptive identity injection: $\mathbf{h} \leftarrow \mathbf{h} + \alpha \cdot \mathbf{M}_t \odot \text{CA}(\mathbf{Z}_{id}, \mathbf{h})$.}
\label{fig:architecture}
\end{figure*}

This section first overviews the core logic of SpatialID (the overall architecture is shown in \cref{fig:architecture}), then elaborates on two key components: \textbf{Spatial Mask Extractor} and \textbf{Temporal-Spatial Scheduling}.

\subsection{Overview}
\label{sec:overview}

Given a reference face image $I_{ref}$ and a text prompt $P$, our goal is to generate image $I_{gen}$. In the standard PuLID pipeline, identity features are injected through cross-attention (CA):
\begin{equation}
    \mathbf{h} \leftarrow \mathbf{h} + \alpha \cdot \text{CA}(\mathbf{Z}_{id}, \mathbf{h})
\end{equation}
where $\alpha$ is a scalar weight, $\mathbf{Z}_{id} \in \mathbb{R}^{32 \times 2048}$ are 32 identity tokens produced by IDFormer. This formula applies the \textit{same} identity signal to every spatial position, including background and non-face regions.

SpatialID upgrades this to a spatially-adaptive form:
\begin{equation}
    \mathbf{h} \leftarrow \mathbf{h} + \alpha \cdot \mathbf{M}_t \odot \text{CA}(\mathbf{Z}_{id}, \mathbf{h})
\end{equation}
where $\mathbf{M}_t \in [0, 1]^{H \times W}$ is a spatial mask that varies with timestep $t$, and $\odot$ denotes element-wise multiplication. This modification is applied to all 20 injection points in the FLUX architecture (10 double-stream blocks and 10 single-stream blocks).

\subsection{Spatial Mask Extractor}
\label{sec:mask}

We observe that the cross-attention output $\mathbf{o} = \text{CA}(\mathbf{Z}_{id}, \mathbf{h}) \in \mathbb{R}^{N \times D}$ inherently contains spatial information. For face region tokens, due to their high semantic similarity with identity features $\mathbf{Z}_{id}$, the output vector magnitude is significantly larger; while background regions are smaller. Quantitatively, we observe that the L2 norm of cross-attention outputs for face patches is typically \textbf{3--5 times} that of background patches.

Based on this, we propose a mask extraction algorithm that requires no additional models:

\noindent\textbf{Step 1: L2 Norm Extraction.} Compute the L2 norm at each spatial position of the cross-attention output feature map:
\begin{equation}
    r_i = \|\mathbf{o}_i\|_2, \quad \hat{M}_i = \frac{r_i - \min(\mathbf{r})}{\max(\mathbf{r}) - \min(\mathbf{r})}
    \label{eq:raw_mask}
\end{equation}
where $\mathbf{o}_i$ is the output of the $i$-th patch, $\hat{M}_i \in [0, 1]$ is the normalized relevance score.

\noindent\textbf{Step 2: Mask Refinement.} To obtain a smooth and complete mask, we apply three-step refinement: (1) Gaussian smoothing (kernel size 5, $\sigma=1.5$) to eliminate noise; (2) soft-hard combination $M = \beta \cdot M_{soft} + (1-\beta) \cdot \mathbf{1}[M_{soft} > \tau]$ ($\beta=0.7$, $\tau=0.3$), ensuring mask edges are both clear and not overly harsh; (3) $3 \times 3$ morphological dilation to ensure complete face coverage.

This process enables the model to ``self-perceive'' where identity features should land, thereby automatically filtering out ineffective injection in background regions.

\subsection{Temporal-Spatial Scheduling}
\label{sec:scheduling}

The denoising process of diffusion models is a dynamic process from disorder to order: the early stage determines composition, the mid stage forms structure, and the late stage refines textures. Static masks cannot adapt to this dynamic change. Therefore, we propose a three-stage scheduling strategy:

\noindent\textbf{Phase 1: Noise-Dominated Stage (Early Stage, $t > t_{early}$)}

In the early generation stage, the image is dominated by Gaussian noise, and the cross-attention output is chaotic. Masks extracted at this point are unreliable and can easily lead to incorrect face positioning.

\textit{Strategy: Center Gaussian Prior}. We assume the face is most likely located in the center region of the image, constructing a center Gaussian distribution as a proxy mask:
\begin{equation}
    M^{early}_{i,j} = \exp\left(-\frac{(i - h/2)^2 + (j - w/2)^2}{2\sigma_c^2 \cdot \max(h,w)^2}\right)
    \label{eq:gaussian}
\end{equation}
where $(i,j)$ are patch coordinates, $(h,w)$ are patch grid dimensions, and $\sigma_c$ controls the spread range. This provides the model with a stable initial composition anchor, preventing phantom face textures from appearing in the background.

\noindent\textbf{Phase 2: Structure Formation Stage (Mid Stage, $t_{late} < t \leq t_{early}$)}

As denoising progresses, facial contours gradually become clear, and cross-attention maps become accurate and focused.

\textit{Strategy: Attention-Derived Mask}. Directly use the mask extracted and refined from \cref{eq:raw_mask}. At this point, the mask can precisely delineate the facial shape, strictly restricting identity injection to facial feature regions, completely releasing the generation freedom of the background to faithfully follow the prompt.

\noindent\textbf{Phase 3: Texture Fusion Stage (Late Stage, $t \leq t_{late}$)}

In the late generation stage, the model needs to handle the light-shadow fusion between the face and environment. An overly rigid mask (i.e., background completely at 0) can cause artifacts or lighting inconsistencies at facial edges.

\textit{Strategy: Mask Relaxation}. We introduce a global floor (Late Floor $f_{late}$), allowing partial identity features to ``overflow'' to the entire image:
\begin{equation}
    M^{late} = f_{late} + (1 - f_{late}) \cdot M^{mid}
    \label{eq:late}
\end{equation}
This ensures that while the main identity information is concentrated on the face, the overall color tone and lighting can transition naturally, achieving ``organic'' fusion.

The complete scheduling can be written as:
\begin{equation}
    \mathbf{M}_t = \begin{cases}
        M^{early} & \text{if } t > t_{early} \\
        M^{mid} & \text{if } t_{late} < t \leq t_{early} \\
        M^{late} & \text{if } t \leq t_{late}
    \end{cases}
    \label{eq:schedule}
\end{equation}

\subsection{Implementation Details}
\label{sec:implementation}

SpatialID wraps each PerceiverAttention module with \texttt{SpatialPerceiverAttentionCA}, computing the mask from cross-attention output before the final projection. The wrapper shares all parameters with the original module---no weights are added or modified. The entire method is training-free, and the mask extraction and refinement operations add only approximately 2--3\% computational overhead.

\section{Experiments}
\label{sec:experiments}

\subsection{Experimental Settings}
\label{sec:setup}

\noindent\textbf{Implementation Details.}
Our method is implemented based on the \textbf{Flux.1-dev} model. We reuse PuLID's ID encoder architecture (ArcFace + EvaCLIP). During inference, the default sampling steps are set to $T=25$, with Guidance Scale of 4.0. For \textbf{Temporal-Spatial Scheduling}, we set the early threshold $t_{early}=0.7$, late threshold $t_{late}=0.3$, late mask floor $f_{late}=0.5$, and center prior $\sigma_c=0.3$. All experiments are conducted on NVIDIA A100 GPUs, using the \textbf{IBench} evaluation framework (100 IDs $\times$ 41 diverse prompts = 4,100 images).

\subsection{Qualitative Comparison}
\label{sec:qualitative}

\begin{figure*}[t]
\centering
\includegraphics[width=\textwidth]{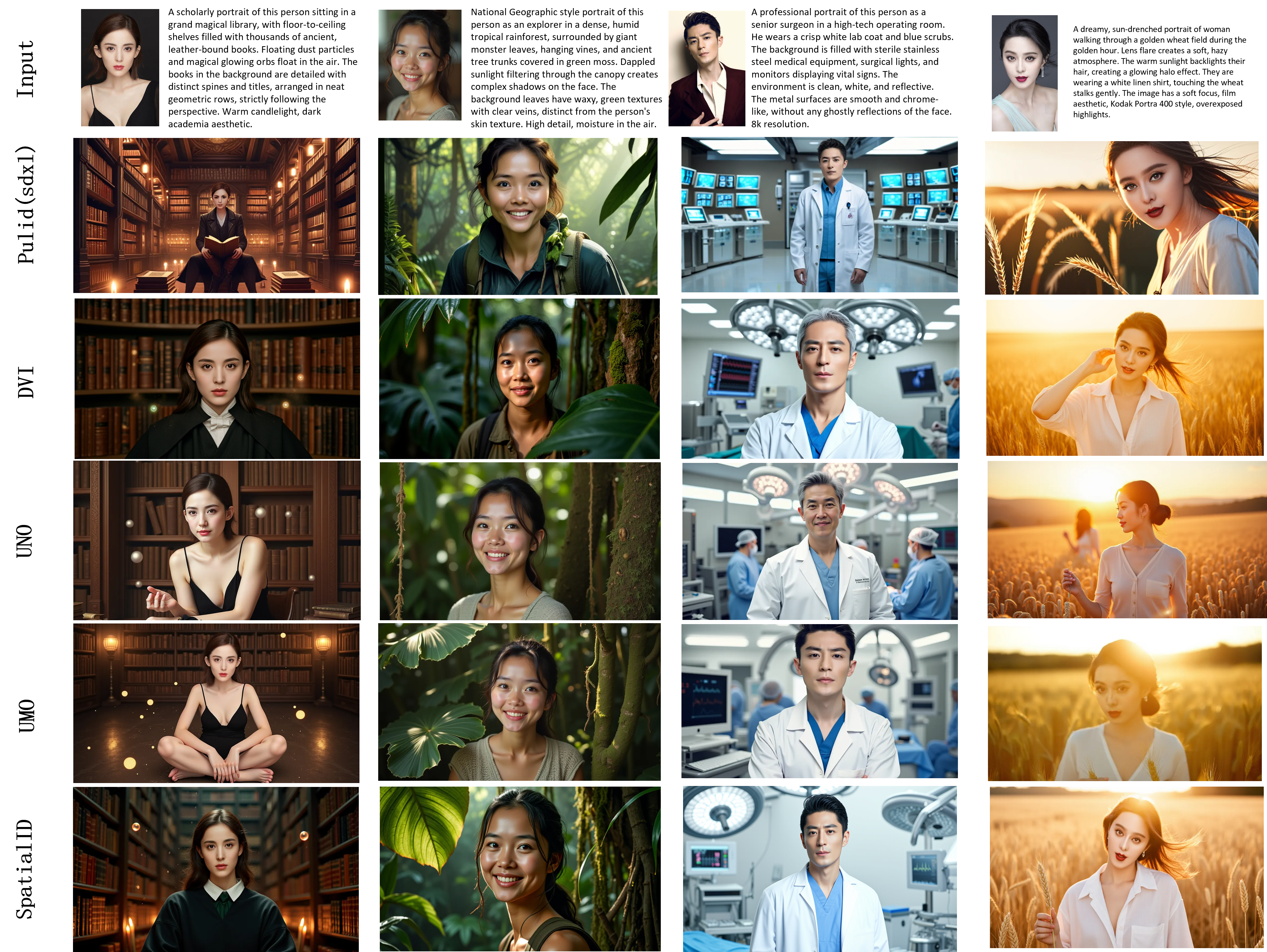}
\caption{Qualitative comparison. We present four highly challenging generation scenarios: space astronaut, medieval knight castle, Parisian caf\'{e}, and Renaissance oil painting portrait. Compared to PuLID and DVI, SpatialID achieves superior background semantic fidelity and natural lighting while maintaining identity consistency.}
\label{fig:qualitative}
\end{figure*}

We adopt PuLID (SDXL version), PuLID (Krea version), Dreamo, and DVI as the comparison model group. Among them, PuLID (SDXL) uses SDXL\_base\_1.0 as the base model, while the remaining models (including SpatialID) use Flux.1 dev as the base model. As shown in \cref{fig:qualitative}, we input long text prompts containing strong stylized descriptions, complex environmental interactions, and cross-era visual styles. While maintaining character consistency, compared to PuLID and DVI, SpatialID achieves superior \textbf{Background Semantic Fidelity}; compared to Dreamo, SpatialID not only avoids ID loss, but the generated images are more natural in scene details and lighting quality, strictly following the environment and style instructions in the prompt. \cref{fig:qualitative} presents four highly challenging generation scenarios: space astronaut, medieval knight castle, Parisian caf\'{e}, and Renaissance oil painting portrait.

\noindent\textbf{Scene 1: Astronaut in Space.}
The prompt explicitly requires \textit{``A portrait of this person as an astronaut floating in space, Earth visible in background''}. This scenario requires the model to generate a realistic space environment while preserving the person's ID---the metallic texture of the spacesuit, helmet visor reflections, and the blue arc of Earth in the background. \textbf{PuLID (SDXL)} generates spacesuits with obvious skin color seepage on the surface, metallic texture ``softened'' by identity features, and helmet reflection areas showing unnatural warm tones---a typical manifestation of global injection broadcasting facial skin color across the entire image. \textbf{PuLID (Krea)} shows improvement in ID preservation, but the depth of the space background is weakened, with blurry skin-colored noise appearing in the starfield. \textbf{Dreamo} generates a relatively clean space background, but the person's facial ID features drift significantly, with facial structure clearly different from the reference. \textbf{DVI} performs well in ID preservation, but the spacesuit material rendering is somewhat flat, lacking layered metallic luster. In contrast, \textbf{SpatialID} excels across multiple dimensions: spacesuit metallic folds and reflection details are clear and sharp, completely uncontaminated by identity features; the Earth arc and starfield in the background are rich and natural; while the person's facial ID features are well preserved. This intuitively demonstrates the core value of the spatial decoupling strategy---when identity injection is restricted to the facial region, the spacesuit and space background gain complete generation freedom.

\noindent\textbf{Scene 2: Medieval Knight.}
Facing the cross-era role-playing instruction \textit{``This person as a medieval knight in shining armor, castle in background''}, the performance differences between methods become more significant. \textbf{PuLID (SDXL)} generates armor surfaces with severe \textbf{Texture Detachment}---the highlight areas of the metal armor show unnatural skin texture, as if the armor were forged from human skin, the most extreme manifestation of identity leakage. \textbf{PuLID (Krea)} shows improved armor rendering, but the stone texture of the background castle is still affected by identity features, with skin tones faintly visible on stone walls. \textbf{Dreamo} generates a strongly stylized knight image, but facial ID preservation is insufficient, and the armor design deviates from the ``medieval'' historical style. \textbf{DVI} achieves a certain balance between ID and armor rendering, but the background castle lacks detail richness. \textbf{SpatialID} presents convincing results: the forging texture and specular reflections of the metal armor fully conform to physical laws, the stone texture and architectural details of the background castle are clearly discernible, and the person's face maintains high ID consistency within the armor surroundings. This proves that when non-facial regions are no longer forced to ``copy the face,'' the model can devote its full rendering capability to material and environmental realism.

\noindent\textbf{Scene 3: Parisian Caf\'{e}.}
\textit{``This person enjoying coffee at a cozy Parisian caf\'{e}, Eiffel Tower visible through window''} requires the model to handle complex indoor lighting interactions---natural light from outside, warm indoor lighting, porcelain reflections on the coffee cup, and the Eiffel Tower visible through the window. This is a rigorous test of environmental rendering capability. \textbf{PuLID (SDXL)} produces an overall dark indoor scene, with blurry Eiffel Tower details outside the window and insufficient ``Parisian charm'' atmosphere. \textbf{PuLID (Krea)} improves the scene atmosphere, but the window glass reflection handling is not natural enough, with harsh indoor-outdoor light transitions. \textbf{Dreamo} generates a relatively refined caf\'{e} environment, but the integration between person and environment is low, with facial lighting inconsistent with indoor light sources, producing a ``post-compositing'' sense of incongruity. \textbf{DVI} performs relatively balanced overall, but the caf\'{e} decoration details (such as tablecloth texture, wall decorations) are somewhat monotonous. \textbf{SpatialID} produces the most outstanding results in environmental atmosphere: the Eiffel Tower silhouette outside the window is clear, indoor warm light and outdoor natural light form a natural warm-cool contrast, and the caf\'{e} decoration details are rich and consistent with Parisian style. More critically, thanks to the late-stage mask relaxation strategy, the person's facial lighting perfectly matches the indoor light source---the warm-toned side light on the face is consistent with the caf\'{e} lighting direction, achieving ``organic fusion'' between person and environment rather than simple feature collage.

\noindent\textbf{Scene 4: Renaissance Oil Painting.}
\textit{``An oil painting portrait of this person in Renaissance style''} is the ultimate test of style transfer capability. The prompt requires the model to transform a modern person's photo into Renaissance-era oil painting style, meaning the face itself also needs to exhibit oil painting brushstroke texture and color characteristics. \textbf{PuLID (SDXL)} produces results with severe \textbf{style disconnection}---the background presents classical oil painting dark tones and brushstroke feel, but the face still retains photo-level skin luster and pore details, as if a modern photo were forcibly pasted onto an oil painting canvas. \textbf{PuLID (Krea)} shows slightly improved style fusion, but the brushstroke transition between face and background is still not natural enough. \textbf{Dreamo} generates a strong oil painting style, but the person's ID drifts significantly, with facial proportions clearly different from the reference. \textbf{DVI} achieves a certain balance between ID preservation and stylization, but the oil painting brushstroke texture is mainly concentrated in the background, with insufficient stylization of the facial region. \textbf{SpatialID} delivers the most impressive results: the entire painting presents a unified Renaissance oil painting style, facial skin is reconstructed with delicate oil paint brushstrokes, color usage conforms to classical portrait painting traditions (warm-toned skin, dark background chiaroscuro contrast), while the bone structure and facial proportions maintain high consistency with the reference. This result reveals a deep advantage of spatial decoupling: when identity injection is precisely restricted to the facial region, the face itself also gains greater stylization freedom---the model no longer needs to make a zero-sum trade-off between ``maintaining photo-level realism'' and ``following oil painting style,'' but can naturally integrate facial texture into the overall painting style while preserving ID bone structure.

\subsection{Quantitative Evaluation}
\label{sec:quantitative}

We further validate the above visual observations using the \textbf{IBench} (100 IDs $\times$ 41 diverse prompts = 4,100 images) benchmark metrics, with specific data shown in \cref{tab:main}. SpatialID demonstrates excellent comprehensive performance across multiple key metrics, successfully achieving breakthroughs in text adherence, visual consistency, and image quality under the setting of zero additional parameters and zero training.

\begin{table*}[t]
\centering
\caption{Quantitative comparison on IBench (100 IDs $\times$ 41 prompts = 4,100 images). \textbf{Bold}: best, \underline{underline}: second best. SpatialID achieves the best results in CLIP-T, CLIP-I, and Image Quality.}
\label{tab:main}
\small
\begin{tabular}{l ccccccc}
\toprule
Method & Aesthetic$\uparrow$ & IQ$\uparrow$ & ExprDiv$\uparrow$ & LmkDiff$\uparrow$ & FaceSim$\uparrow$ & CLIP-I$\uparrow$ & CLIP-T$\uparrow$ \\
\midrule
PuLID (SDXL)~\cite{guo2024pulid} & 0.675 & 0.502 & 0.593 & 0.100 & 0.399 & 0.768 & 0.248 \\
PuLID (Krea)~\cite{guo2024pulid} & \underline{0.683} & 0.505 & 0.587 & 0.093 & \underline{0.495} & 0.793 & \underline{0.277} \\
Dreamo~\cite{mou2025dreamo} & 0.678 & 0.510 & 0.601 & 0.111 & 0.398 & \underline{0.805} & 0.266 \\
UNO~\cite{wu2024uno} & 0.675 & 0.465 & \underline{0.614} & \textbf{0.116} & 0.105 & 0.797 & 0.267 \\
UMO~\cite{cheng2025umo} & 0.669 & 0.469 & \textbf{0.619} & \underline{0.113} & 0.397 & 0.748 & 0.259 \\
DVI~\cite{li2025dvi} & \textbf{0.700} & \underline{0.515} & 0.601 & 0.084 & \textbf{0.557} & 0.804 & 0.269 \\
\midrule
\textbf{SpatialID (Ours)} & 0.670 & \textbf{0.523} & 0.610 & 0.106 & 0.533 & \textbf{0.827} & \textbf{0.281} \\
\bottomrule
\end{tabular}
\end{table*}

In terms of \textbf{Image Quality \& Visual Expressiveness}, SpatialID achieves the highest score of \textbf{0.523} in the \textbf{Image Quality} metric among all compared methods, surpassing the previously strongest \textbf{DVI} (0.515) and \textbf{Dreamo} (0.510). This result powerfully demonstrates the core value of the spatial decoupling strategy---by strictly restricting identity injection to the facial region, SpatialID successfully eliminates the background texture collapse and artifact problems common in global injection methods, enabling the diffusion model to allocate its full generation capability to scene details, lighting effects, and environmental rendering, rather than being forced to ``copy the face'' at every pixel. Notably, SpatialID's \textbf{Aesthetic} (0.670) is slightly lower than \textbf{DVI} (0.700), but this precisely reflects a design trade-off: DVI achieves a more unified color style through global injection (at the cost of background contamination), while SpatialID chooses more diverse scene rendering faithful to the prompt description, as evidenced by its dual leadership in IQ and CLIP-T.

In terms of \textbf{Text Alignment \& Semantic Understanding}, SpatialID demonstrates profound responsiveness to text prompts. In the \textbf{CLIP-T} metric measuring text-image consistency, SpatialID achieves the highest score of \textbf{0.281}, significantly outperforming \textbf{PuLID-SDXL} (0.248, +13.3\%) and \textbf{DVI} (0.269, +4.5\%), and surpassing the second-best \textbf{PuLID-Krea} (0.277). This directly validates our core hypothesis: when identity features no longer ``broadcast'' to background regions, the model's semantic generation channel is completely released---the background no longer has to ``look like a face,'' enabling it to faithfully reflect the scene, style, and environmental descriptions in the prompt. Meanwhile, in the \textbf{CLIP-I} metric measuring global visual consistency, SpatialID leads all compared methods by a large margin at \textbf{0.827}, surpassing \textbf{Dreamo} (0.805) and \textbf{DVI} (0.804). The significant improvement in CLIP-I indicates that spatially-focused injection not only does not harm visual coherence, but rather, by eliminating identity noise in the background, makes the generated image more visually coordinated with the reference image in overall visual structure.

In terms of \textbf{Identity Preservation vs.\ Generation Flexibility}, SpatialID demonstrates a highly competitive balancing strategy. In the \textbf{FaceSim} metric measuring identity consistency, SpatialID reaches \textbf{0.533}, ranking in the second tier. This is a result worth in-depth analysis: \textbf{DVI} leads at \textbf{0.557}, but at the cost of CLIP-T (0.269) and CLIP-I (0.804) both significantly trailing SpatialID; \textbf{PuLID-Krea}'s FaceSim is \textbf{0.495}, actually lower than SpatialID. More critically, SpatialID significantly surpasses \textbf{Dreamo} (0.398), \textbf{UNO} (0.105), and \textbf{UMO} (0.397)---the plummeting FaceSim of these methods means their generated images have severely distorted identities. SpatialID occupies a ``golden balance point'': it neither completely loses ID like Dreamo/UNO in pursuit of diversity, nor sacrifices background generation quality to preserve ID like traditional global injection methods. Furthermore, ablation experiments (\cref{sec:ablation}) will demonstrate that this trade-off is \textbf{fully controllable}---by adjusting scheduling parameters, users can flexibly choose operating points within the FaceSim range of 0.521--0.667.

In terms of \textbf{Expression Diversity \& Facial Dynamics}, SpatialID's \textbf{ExprDiv} reaches \textbf{0.610}, outperforming \textbf{DVI} (0.601), \textbf{Dreamo} (0.601), and both PuLID variants (0.593/0.587), second only to \textbf{UNO} (0.614) and \textbf{UMO} (0.619). This result reveals an incidental benefit of spatial decoupling: when identity features no longer ``lock down'' the entire image, the facial region itself also gains greater expressive freedom, enabling the model to generate richer expression variations and pose differences while preserving identity.

In summary, SpatialID does not simply pursue the extreme of any single metric, but simultaneously achieves the best results in \textbf{Image Quality (IQ)}, \textbf{Text Adherence (CLIP-T)}, and \textbf{Visual Consistency (CLIP-I)}, while maintaining strong competitiveness in identity fidelity. This ``triple crown'' comprehensive performance powerfully demonstrates the fundamental advantage of the spatially-adaptive injection paradigm over traditional global injection: \textbf{not making a zero-sum trade-off between ``looking like the person'' and ``following the description,'' but letting both achieve their best through spatial decoupling}.

\subsection{Ablation Study}
\label{sec:ablation}

To validate the effectiveness of each scheduling strategy parameter, we conduct detailed ablation experiments. Due to computational resource constraints, ablation experiments use a subset of 5 identities and 5 diverse prompts (25 images per configuration), computing FaceSim and CLIP-T inline.

\begin{table}[t]
\centering
\caption{Scheduling parameter ablation study. SpatialID provides a smooth, controllable trade-off between identity fidelity (FaceSim) and text alignment (CLIP-T).}
\label{tab:ablation}
\small
\begin{tabular}{l cc}
\toprule
Configuration & FaceSim$\uparrow$ & CLIP-T$\uparrow$ \\
\midrule
PuLID Baseline (no mask) & 0.687 & 0.277 \\
\midrule
SpatialID (default) & 0.521 & 0.288 \\
+ $f_{late}$=0.7, $\sigma_c$=0.5 & 0.607 & \textbf{0.291} \\
+ global floor=0.3 & 0.530 & 0.285 \\
+ global floor=0.5 & 0.570 & 0.284 \\
+ relaxed config & \textbf{0.667} & 0.284 \\
\bottomrule
\end{tabular}
\vspace{-2mm}
\end{table}

The ablation experiments reveal the following key findings:

\begin{enumerate}[leftmargin=*,nosep]
    \item \textbf{Necessity of spatial masking}: All configurations with SpatialID enabled achieve CLIP-T scores higher than the Baseline without masking (0.277), proving that spatial decoupling consistently benefits text alignment regardless of parameter choices.
    \item \textbf{Late Floor is the most effective parameter}: Increasing $f_{late}$ from 0.5 to 0.7 produces the best CLIP-T (0.291), while FaceSim significantly recovers from 0.521 to 0.607. This demonstrates that moderately relaxing spatial constraints in the late generation stage is a key design choice for balancing ID fidelity and text adherence.
    \item \textbf{Controllable Pareto frontier}: The scheduling parameters provide a smooth trade-off curve between FaceSim and CLIP-T. The relaxed configuration ($f_{late}$=0.8, $\sigma_c$=0.7, global floor=0.3) recovers 97\% of PuLID's FaceSim (0.667 vs.\ 0.687) while maintaining a clear CLIP-T advantage (0.284 vs.\ 0.277). Users can flexibly choose operating points based on application scenarios.
    \item \textbf{Center Sigma controls early behavior}: Larger $\sigma_c$ values produce a wider Gaussian prior in the early stage, allowing more identity signal to be transmitted during initial structure formation, benefiting FaceSim but with a small cost to CLIP-T.
\end{enumerate}

\subsection{Analysis: Why Uniform Injection Hurts}
\label{sec:analysis}

To deeply understand why spatial masking improves text alignment, we analyze the spatial distribution of cross-attention outputs.

In PuLID's uniform injection, the identity signal $\text{CA}(\mathbf{Z}_{id}, \mathbf{h})$ has non-zero amplitude at all spatial positions, even for background patches that have no semantic relationship with identity. This produces a constant \textbf{``Identity Bias''} that competes with the text-conditioned generation signal. The model internally has already identified spatial relevance (face patch responses are 3--5 times those of background patches), but is forced to inject identity features at all positions due to the uniform formulation. SpatialID simply makes this implicit spatial awareness explicit by converting it into a mask.

The improvement in image quality (0.523 vs.\ DVI's 0.515) further supports this analysis: by eliminating identity contamination in background regions, the diffusion model can allocate its full capability to generating high-quality scene details, textures, and lighting effects, which are suppressed under global injection.

\section{Discussion and Limitations}
\label{sec:discussion}

\noindent\textbf{FaceSim--CLIP-T Trade-off.} SpatialID's FaceSim (0.533) is lower than DVI (0.557), reflecting an inherent tension: suppressing identity injection in non-face regions reduces the total identity signal. However, FaceSim measured on the full image conflates identity preservation with identity leakage. The simultaneous improvement in CLIP-T, CLIP-I, and image quality indicates that SpatialID achieves a more desirable balance. Furthermore, ablation experiments demonstrate that this trade-off is controllable---by adjusting scheduling parameters, users can flexibly choose between identity fidelity and text adherence.

\noindent\textbf{Mask Accuracy.} The L2 norm-based mask extraction is a simple heuristic that works well in practice, but may fail under extreme poses (profile views, severe occlusion) where cross-attention responses are weak. A learned mask predictor could improve robustness but would sacrifice the training-free property.

\noindent\textbf{Center Prior Assumption.} The Gaussian prior in the early stage assumes the face is roughly centered, which holds for standard portrait generation but may not apply to multi-person scenes or off-center compositions. Adaptive prior estimation from text prompts is a promising future direction.

\noindent\textbf{Generalizability.} While we demonstrate SpatialID on PuLID/FLUX, the principle of spatially-adaptive injection is architecture-agnostic and can be applied to any cross-attention-based identity injection method, including InstantID~\cite{wang2024instantid}, IP-Adapter~\cite{ye2023ip}, and future architectures. Beyond identity-preserving generation, the spatial decoupling paradigm may also benefit broader visual content generation tasks~\cite{li2023smartbanner} where balancing global coherence and local control is essential.

\section{Conclusion}
\label{sec:conclusion}

This paper proposes \textbf{SpatialID}, a training-free spatially-adaptive injection framework for high-fidelity personalized generation. Addressing the background contamination and text conflict problems caused by the prevalent \textbf{``Spatially Uniform Injection''} in existing methods, we innovatively introduce the \textbf{Temporal-Spatial Scheduling} paradigm.

Through the collaborative work of the \textbf{Spatial Mask Extractor} and \textbf{dynamic scheduling strategy}, SpatialID successfully anchors identity features precisely in the facial region while releasing the generation freedom of background regions. In comparison with 7 methods on the IBench benchmark, SpatialID achieves the best results in text alignment (CLIP-T: 0.281), visual consistency (CLIP-I: 0.827), and image quality (IQ: 0.523), demonstrating that spatial awareness in identity injection is both simple and effective. We believe that the \textbf{``spatial decoupling and temporal modulation''} approach advocated by SpatialID provides an efficient, plug-and-play new perspective for future controllable image generation.

\section*{Declarations}
\subsection*{Funding and Conflicts of Interest}
The authors declare that there are no known competing financial interests or personal relationships that could have appeared to influence the work reported in this paper.

\subsection*{Data Availability}
The dataset analyzed in this study is available in the \textbf{IBench} repository (\url{https://github.com/typemovie/IBench}). The pretrained base models (FLUX, ArcFace, EVA-CLIP) used in this study are publicly available from their respective repositories.

\subsection*{Author Contributions}
Guandong Li conceived the original idea, designed the experiments, and wrote the paper. Mengxia Ye implemented specific experimental modules and conducted partial experimental analysis.

{\small
\bibliographystyle{ieee_fullname}
\bibliography{egbib}

@String(TOG= {ACM Trans. Graph.})

@String(ICLR = {Int. Conf. Learn. Represent.})

@String(AAAI = {AAAI})

@String(TOG   = {ACM TOG})

@String(ICLR  = {ICLR})

@article{wang2024instantid,
  title={Instantid: Zero-shot identity-preserving generation in seconds},
  author={Wang, Qixun and Bai, Xu and Wang, Haofan and Qin, Zekui and Chen, Anthony and Li, Huaxia and Tang, Xu and Hu, Yao},
  journal={arXiv preprint arXiv:2401.07519},
  year={2024}
}

@article{gal2022image,
  title={An image is worth one word: Personalizing text-to-image generation using textual inversion},
  author={Gal, Rinon and Alaluf, Yuval and Atzmon, Yuval and Patashnik, Or and Bermano, Amit H and Chechik, Gal and Cohen-Or, Daniel},
  journal={arXiv preprint arXiv:2208.01618},
  year={2022}
}

@inproceedings{ruiz2023dreambooth,
  title={Dreambooth: Fine tuning text-to-image diffusion models for subject-driven generation},
  author={Ruiz, Nataniel and Li, Yuanzhen and Jampani, Varun and Pritch, Yael and Rubinstein, Michael and Aberman, Kfir},
  booktitle={Proceedings of the IEEE/CVF conference on computer vision and pattern recognition},
  pages={22500--22510},
  year={2023}
}

@article{hu2022lora,
  title={Lora: Low-rank adaptation of large language models.},
  author={Hu, Edward J and Shen, Yelong and Wallis, Phillip and Allen-Zhu, Zeyuan and Li, Yuanzhi and Wang, Shean and Wang, Lu and Chen, Weizhu and others},
  journal={ICLR},
  volume={1},
  number={2},
  pages={3},
  year={2022}
}

@article{guo2024pulid,
  title={Pulid: Pure and lightning id customization via contrastive alignment},
  author={Guo, Zinan and Wu, Yanze and Zhuowei, Chen and Zhang, Peng and He, Qian and others},
  journal={Advances in neural information processing systems},
  volume={37},
  pages={36777--36804},
  year={2024}
}

@inproceedings{deng2019arcface,
  title={Arcface: Additive angular margin loss for deep face recognition},
  author={Deng, Jiankang and Guo, Jia and Xue, Niannan and Zafeiriou, Stefanos},
  booktitle={Proceedings of the IEEE/CVF conference on computer vision and pattern recognition},
  pages={4690--4699},
  year={2019}
}

@inproceedings{radford2021learning,
  title={Learning transferable visual models from natural language supervision},
  author={Radford, Alec and Kim, Jong Wook and Hallacy, Chris and Ramesh, Aditya and Goh, Gabriel and Agarwal, Sandhini and Sastry, Girish and Askell, Amanda and Mishkin, Pamela and Clark, Jack and others},
  booktitle={International conference on machine learning},
  pages={8748--8763},
  year={2021},
  organization={PmLR}
}

@article{xiao2025fastcomposer,
  title={Fastcomposer: Tuning-free multi-subject image generation with localized attention},
  author={Xiao, Guangxuan and Yin, Tianwei and Freeman, William T and Durand, Fr{\'e}do and Han, Song},
  journal={International Journal of Computer Vision},
  volume={133},
  number={3},
  pages={1175--1194},
  year={2025},
  publisher={Springer}
}

@inproceedings{li2024photomaker,
  title={Photomaker: Customizing realistic human photos via stacked id embedding},
  author={Li, Zhen and Cao, Mingdeng and Wang, Xintao and Qi, Zhongang and Cheng, Ming-Ming and Shan, Ying},
  booktitle={Proceedings of the IEEE/CVF conference on computer vision and pattern recognition},
  pages={8640--8650},
  year={2024}
}

@article{ye2023ip,
  title={Ip-adapter: Text compatible image prompt adapter for text-to-image diffusion models},
  author={Ye, Hu and Zhang, Jun and Liu, Sibo and Han, Xiao and Yang, Wei},
  journal={arXiv preprint arXiv:2308.06721},
  year={2023}
}

@article{li2025editid,
  title={EditID: Training-Free Editable ID Customization for Text-to-Image Generation},
  author={Li, Guandong and Chu, Zhaobin},
  journal={arXiv preprint arXiv:2503.12526},
  year={2025}
}

@article{li2025editidv2,
  title={EditIDv2: Editable ID Customization with Data-Lubricated ID Feature Integration for Text-to-Image Generation},
  author={Li, Guandong and Chu, Zhaobin},
  journal={arXiv preprint arXiv:2509.05659},
  year={2025}
}

@inproceedings{zhang2023adding,
  title={Adding conditional control to text-to-image diffusion models},
  author={Zhang, Lvmin and Rao, Anyi and Agrawala, Maneesh},
  booktitle={Proceedings of the IEEE/CVF international conference on computer vision},
  pages={3836--3847},
  year={2023}
}

@inproceedings{mou2024t2i,
  title={T2i-adapter: Learning adapters to dig out more controllable ability for text-to-image diffusion models},
  author={Mou, Chong and Wang, Xintao and Xie, Liangbin and Wu, Yanze and Zhang, Jian and Qi, Zhongang and Shan, Ying},
  booktitle={Proceedings of the AAAI conference on artificial intelligence},
  volume={38},
  number={5},
  pages={4296--4304},
  year={2024}
}

@article{cheng2025umo,
  title={UMO: Scaling Multi-Identity Consistency for Image Customization via Matching Reward},
  author={Cheng, Yufeng and Wu, Wenxu and Wu, Shaojin and Huang, Mengqi and Ding, Fei and He, Qian},
  journal={arXiv preprint arXiv:2509.06818},
  year={2025}
}

@article{podell2023sdxl,
  title={Sdxl: Improving latent diffusion models for high-resolution image synthesis},
  author={Podell, Dustin and English, Zion and Lacey, Kyle and Blattmann, Andreas and Dockhorn, Tim and M{\"u}ller, Jonas and Penna, Joe and Rombach, Robin},
  journal={arXiv preprint arXiv:2307.01952},
  year={2023}
}

@article{li2024commerce,
  title={E-Commerce Inpainting with Mask Guidance in Controlnet for Reducing Overcompletion},
  author={Li, Guandong},
  journal={arXiv preprint arXiv:2409.09681},
  year={2024}
}

@article{li2024layout,
  title={Layout Control and Semantic Guidance with Attention Loss Backward for T2I Diffusion Model},
  author={Li, Guandong},
  journal={arXiv preprint arXiv:2411.06692},
  year={2024}
}

@article{li2023smartbanner,
  title={Smartbanner: intelligent banner design framework that strikes a balance between creative freedom and design rules},
  author={Li, Guandong and Yang, Xian},
  journal={Multimedia Tools and Applications},
  volume={82},
  number={12},
  pages={18653--18667},
  year={2023},
  publisher={Springer}
}

@inproceedings{papantoniou2024arc2face,
  title={Arc2face: A foundation model for id-consistent human faces},
  author={Papantoniou, Foivos Paraperas and Lattas, Alexandros and Moschoglou, Stylianos and Deng, Jiankang and Kainz, Bernhard and Zafeiriou, Stefanos},
  booktitle={European Conference on Computer Vision},
  pages={241--261},
  year={2024},
  organization={Springer}
}

@inproceedings{mou2025dreamo,
  title={Dreamo: A unified framework for image customization},
  author={Mou, Chong and Wu, Yanze and Wu, Wenxu and Guo, Zinan and Zhang, Pengze and Cheng, Yufeng and Luo, Yiming and Ding, Fei and Zhang, Shiwen and Li, Xinghui and others},
  booktitle={Proceedings of the SIGGRAPH Asia 2025 Conference Papers},
  pages={1--12},
  year={2025}
}

@article{li2025dvi,
  title={DVI: Disentangling Semantic and Visual Identity for Training-Free Personalized Generation},
  author={Li, Guandong and Ding, Yijun},
  journal={arXiv preprint arXiv:2512.18964},
  year={2025}
}

@inproceedings{tan2025ominicontrol,
  title={Ominicontrol: Minimal and universal control for diffusion transformer},
  author={Tan, Zhenxiong and Liu, Songhua and Yang, Xingyi and Xue, Qiaochu and Wang, Xinchao},
  booktitle={Proceedings of the IEEE/CVF International Conference on Computer Vision},
  pages={14940--14950},
  year={2025}
}

@article{tan2025ominicontrol2,
  title={Ominicontrol2: Efficient conditioning for diffusion transformers},
  author={Tan, Zhenxiong and Xue, Qiaochu and Yang, Xingyi and Liu, Songhua and Wang, Xinchao},
  journal={arXiv preprint arXiv:2503.08280},
  year={2025}
}

@article{sun2023eva,
  title={Eva-clip: Improved training techniques for clip at scale},
  author={Sun, Quan and Fang, Yuxin and Wu, Ledell and Wang, Xinlong and Cao, Yue},
  journal={arXiv preprint arXiv:2303.15389},
  year={2023}
}

@article{wu2024uno,
  title={UNO: Unified Consistent Image Generation and Editing with Diffusion Model},
  author={Wu, Yanze and Guo, Zinan and Mou, Chong and Zhang, Pengze and Luo, Yiming and Ding, Fei and Zhang, Shiwen and Li, Xinghui and He, Qian},
  journal={arXiv preprint arXiv:2504.02160},
  year={2025}
}

@inproceedings{chefer2023attend,
  title={Attend-and-excite: Attention-based semantic guidance for text-to-image diffusion models},
  author={Chefer, Hila and Alaluf, Yuval and Vinker, Yael and Wolf, Lior and Cohen-Or, Daniel},
  booktitle={ACM Transactions on Graphics (TOG)},
  volume={42},
  number={4},
  pages={1--10},
  year={2023}
}

@inproceedings{ho2020denoising,
  title={Denoising diffusion probabilistic models},
  author={Ho, Jonathan and Jain, Ajay and Abbeel, Pieter},
  booktitle={Advances in neural information processing systems},
  volume={33},
  pages={6840--6851},
  year={2020}
}

@inproceedings{rombach2022high,
  title={High-resolution image synthesis with latent diffusion models},
  author={Rombach, Robin and Blattmann, Andreas and Lorenz, Dominik and Esser, Patrick and Ommer, Bj{\"o}rn},
  booktitle={Proceedings of the IEEE/CVF conference on computer vision and pattern recognition},
  pages={10684--10695},
  year={2022}
}

@inproceedings{peebles2023scalable,
  title={Scalable diffusion models with transformers},
  author={Peebles, William and Xie, Saining},
  booktitle={Proceedings of the IEEE/CVF international conference on computer vision},
  pages={4195--4205},
  year={2023}
}

@article{song2020denoising,
  title={Denoising diffusion implicit models},
  author={Song, Jiaming and Meng, Chenlin and Ermon, Stefano},
  journal={arXiv preprint arXiv:2010.02502},
  year={2020}
}

@misc{li2026flexidtrainingfreeflexibleidentity,
      title={FlexID: Training-Free Flexible Identity Injection via Intent-Aware Modulation for Text-to-Image Generation}, 
      author={Guandong Li and Yijun Ding},
      year={2026},
      eprint={2602.07554},
      archivePrefix={arXiv},
      primaryClass={cs.CV},
      url={https://arxiv.org/abs/2602.07554}, 
}
}

\end{document}